# Self-supervised Speech Representations Still Struggle with African American Vernacular English

*Kalvin Chang*[1,*], *Yi-Hui Chou*[1,*], *Jiatong Shi*[1], *Hsuan-Ming Chen*[1],
*Nicole Holliday*[3], *Odette Scharenborg*[2], *David R. Mortensen*[1]

[1]Carnegie Mellon University [2]Delft University of Technology [3]University of California, Berkeley

{kalvinc, yihuic, hsuanmic}@alumni.cmu.edu, {jiatongs, dmortens}@cs.cmu.edu,
nicole.holliday@berkeley.edu, O.E.scharenborg@tudelft.nl

## Abstract

Underperformance of ASR systems for speakers of African American Vernacular English (AAVE) and other marginalized language varieties is a well-documented phenomenon, and one that reinforces the stigmatization of these varieties. We investigate whether or not the recent wave of Self-Supervised Learning (SSL) speech models can close the gap in ASR performance between AAVE and Mainstream American English (MAE). We evaluate four SSL models (wav2vec 2.0, HuBERT, WavLM, and XLS-R) on zero-shot Automatic Speech Recognition (ASR) for these two varieties and find that these models perpetuate the bias in performance against AAVE. Additionally, the models have higher word error rates on utterances with more phonological and morphosyntactic features of AAVE. Despite the success of SSL speech models in improving ASR for low resource varieties, SSL pre-training alone may not bridge the gap between AAVE and MAE.[1]

## 1. Introduction

African-American Vernacular English (AAVE) is a recognized variety of English spoken by many Black Americans that differs phonologically, morphologically, syntactically, and semantically from Mainstream American English (MAE) [1]. This variety has long been stigmatized in public discourses as "incorrect" or "ungrammatical" [2]. Its use may be a proxy for race, as research has indicated that AAVE speakers experience linguistic discrimination. Several experiments conducted via telephone have revealed pervasive biases against AAVE speakers in areas such as housing [3, 4].

The poor performance of ASR on AAVE perpetuates the stigmatization of AAVE [5–7], a representational harm [8]. [9] reported higher Word Error Rates (WER) for Black speakers (0.35) compared to those of white speakers (0.19) on five commercial ASR systems. They ran inference with five ASR systems in a zero-shot fashion on one corpus of AAVE, CORAAL [10], and another corpus of MAE, Voices of California.

Bias against AAVE in ASR causes speakers to feel marginalized and forced to accommodate to MAE norms when using ASR [5, 11]. In [12], elderly Black individuals felt pressured to use "cultural code-switching" to be understood by voice assistants designed to provide important healthcare information. ASR in video interviewing applications may perpetuate discrimination against Black Americans in hiring since the bias against AAVE in ASR may hurt their language scores [6]. Transcribing physician-patient conversations or diagnosing patients using ASR may also preserve racial disparities in healthcare [6]. Bias in ASR performance can thus result in negative social and economic consequences by limiting access to opportunities and services, creating allocational harms [8].

One might assume that approaches that have worked for other low resource languages (LRLs) would work for AAVE. For many LRLs, SSL speech representations (SSLR), such as wav2vec 2.0, HuBERT, and XLS-R, have allowed researchers to leverage unlabeled speech corpora to attain lower WERs and decrease the amount of labeled speech needed [13], attaining 23% WER with 4 hours of recordings in the case of Maori [14].

Wav2vec 2.0 and HuBERT encode granular acoustic [15], phonetic [16], phonological [17, 18], prosodic [19], and dialectal distinctions [20], and as such, have the potential to reduce disparities in WERs between AAVE and MAE, as [9] attributed the disparity in performance primarily to the acoustic model of the ASR system, rather than the language model. In this paper, we find the contrary, contributing:

1. Empirical evidence that SSL pre-training on out of domain data can perpetuate the disparity between AAVE and MAE in zero shot ASR (Section 4.1)
2. Error analysis that reveals SSLRs still struggle with AAVE features (Section 4.2)

## 2. Related Work

Prior studies, including [9], have revealed bias against AAVE in ASR systems. [21] showed significant differences in WER across races and dialects in YouTube's automatic captions. [22] reported low frequencies of common AAVE morphosyntactic features (invariant *be*, perfective *done*, and *ain't*) in the Fisher, Switchboard, LibriSpeech, and TIMIT corpora, suggesting that AAVE speakers are poorly represented in these datasets. Unsurprisingly, Amazon, DeepSpeech, Google, IBM, and Microsoft's ASR systems struggled with these 3 morphosyntactic features. [23] found that AAVE speech with more variable vowel durations leads to higher WER when compared to MAE speech, though rhythmic variation itself does not play a role. In [24], sociophonetic variables explained 20% of errors by Azure Speech on different ethnolects of American English (including AAVE).

Recent research has explored various approaches to improve ASR performance for AAVE. [25] trained a discriminative pronunciation model jointly with a language model, moderately improving ASR for AAVE. [26] demonstrated an 18.6% improvement in models trained exclusively on AAVE compared to those trained on both AAVE and Mainstream American English, indicating that the WER disparity in ASR stems from inadequate representation in training data rather than inherent acoustic challenges posed by AAVE. [27–29] classified AAVE and estimated AAVE dialect density from X-vector, prosodic,

---
* Equal contribution.
[1]Code: https://github.com/cmu-llab/s3m-aave

SSL, and morphosyntactic features with XGBoost. Adding a counterfactual loss term [30] or an equal accuracy ratio regularization term [31] improved ASR performance on CORAAL.

Subsequent studies investigated how data bias affects ASR performance, with [32] illustrating how pre-training data bias, including gender, content, and prosody, affects SSL models in various downstream tasks. [33] found that SSL pre-training with 1 million hours of spontaneous YouTube speech and synthetic speech from Text-to-Speech (TTS) augmentation enhanced recognition of both AAVE and L2 English accents, possibly due to the richer diversity of data on platforms like YouTube. Unlike our study, they do not compare performance on AAVE with that of MAE. Furthermore, [34] employed a dialect classifier to identify AAVE instances in a large corpus for generating pseudo-labeled transcriptions, reducing the WER disparity by a relative 38.5%. Moreover, the Universal Speech Model [35], pre-trained on 12 million hours of unlabeled, multilingual YouTube (spontaneous) data and finetuned on multilingual YouTube and other public data, achieved a state-of-the-art 11.2 WER on the CORAAL dataset of Black speakers, showing that mitigating this type of ASR bias is not insurmountable.

While work on AAVE in ASR has increased recently, it remains unclear whether SSL pre-trained on out-of-domain data (Section 3.2.3) generalizes to AAVE and whether they can close the gap in ASR performance between AAVE and MAE.

## 3. Methodology

Our experimental design follows [9] in that we compare the zero-shot[2] performance of several ASR models on one corpus of AAVE (Section 3.1.1) and another of MAE (Section 3.1.2). We also use propensity score matching to reduce the effects of confounding variables on WER (Section 3.3.1). For ASR, we use an end-to-end architecture with four pre-trained SSL models as features (Section 3.2.2). We trained ASR models with 100 hours of LibriSpeech (Section 3.2.2), which notably favors MAE (due to the low normalized frequencies of 3 common AAVE morphosyntactic features in this corpus [22]). We then examine how the models perform on utterances with more AAVE features (Section 3.3.2).

### 3.1. Datasets

*3.1.1. Corpus of Regional African American Language*

CORAAL [10] is a dataset of audio recordings from over 150 sociolinguistic interviews with individuals who speak AAVE. The interviews were conducted in several cities across the US, including Princeville (a rural, predominantly African American community in North Carolina), Rochester (a mid-sized city in New York), and the District of Columbia (the capital of the U.S.). Personal information such as addresses is redacted, and the data is publicly available. Undergraduate research assistants transcribed the recordings, after which a graduate student in linguistics revised the transcription. Transcriptions preserve morphosyntactic variation (e.g. null copula) but not phonological variation (e.g. velar nasal fronting). Many reduced constructions (e.g. "sposta" for "supposed to") are preserved, but not all are (e.g. "useta" is transcribed as "used to"). AAVE-specific lexical items (e.g. bruh) are also preserved.[3] We use [9]'s 3-city (Princeville, Rochester, and D.C.) subset of CORAAL as well as their segmented utterances, which avoids overlapping speech and pauses. We downsampled from 44.1 kHz to 16 kHz.

---

[2]Finetuning on in-domain AAVE data is an obvious way to mitigate the observed disparity, but our focus is highlighting the disparity itself.

[3]Refer to the CORAAL User Guide for more information on the interview and transcription process.

*3.1.2. Nationwide Speech Project (NSP)*

As a comparison set to CORAAL, we employed the Nationwide Speech Project (NSP) corpus [36]. NSP consists of read and spontaneous speech in Mainstream American English from a homogeneous demographic: young (age 18-25), white speakers with native English-speaking parents. The dataset has balanced gender and regional dialect representation, with 10 speakers from each of [37]'s six dialect regions. The spontaneous speech portion consists of 5 minutes of sociolinguistic interviews per speaker. We obtained the dataset from the authors, as the dataset cannot be distributed without permission. The dataset was recorded in a sound-attenuated booth. Undergraduates transcribed the recordings, and Clopper then reviewed the transcriptions herself. The transcribers annotated overlapping speech, unintelligible words, pauses, paralinguistics, and whispered speech. The transcriptions included filler words and dysfluencies. We downsampled from 44.1 kHz to 16 kHz.

To remove utterances from the interviewer, we extracted word-level timestamps with the Montreal Forced Aligner [38][4] and used the provided speaker labels in the transcript as utterance boundaries for segmentation. We removed all paralinguistic annotations from the transcripts, as well as clips containing overlapping speech and clips shorter than 3 seconds.

Table 1: *Dataset statistics for CORAAL and NSP.* † *refers to statistics after propensity score matching (Section 3.3.1). Average DNS-MOS [39] and UT-MOS [40] scores represent the audio quality on a scale from 1-5.*

| Dataset | Duration (Total) | Duration (Avg) | Age (Avg) | DNS-MOS | UT-MOS |
|---|---|---|---|---|---|
| CORAAL | 19.39 h | 15.71 s | 43.41 | 3.41 | 2.21 |
| NSP | 3.33 h | 12.99 s | 19.18 | 3.66 | 3.00 |
| CORAAL† | 4.02 h | 15.69 s | 25.38 | 3.45 | 2.29 |
| NSP† | 3.33 h | 12.99 s | 19.18 | 3.66 | 3.00 |

*3.1.3. Text Normalization*

[9] removed all punctuation and annotations for overlapping speech, redactions, unintelligibility, inaudible portions, and paralinguistic sounds (e.g. coughing or laughing). They expanded numbers (including dates and ordinal numbers). We revised their text normalization to keep apostrophes, swear words, filler words, and dysfluencies and expanded abbreviations (e.g. "ms" → "miss") to match LibriSpeech. We additionally normalized all the reduced constructions, as LibriSpeech (train_clean_100) does not use reduced forms. Importantly, we applied the above text normalization procedure to both NSP and CORAAL.

### 3.2. ASR models

We trained ASR models on LibriSpeech (train_clean_100) in ESPnet and ran inference in a zero-shot fashion on CORAAL and NSP. The SSL models use a standard Transformer encoder-decoder architecture with SSLRs as features instead of FBANK or MFCC features. The 317 million SSL parameters are frozen during training to see if they can generalize to AAVE. As a baseline, we compare the SSL models with a model with FBANK features.

---

[4]We used the pretrained english_us_arpa American English english_us_arpa acoustic model and pronunciation dictionary.

*3.2.1. Supervised model*

To see if pretrained SSL reduces the disparity between CORAAL and NSP, we compare the performance of SSL on both datasets with a model with FBANK features instead of SSL features. The model follows the same Transformer encoder-decoder architecture and training configuration as the SSL models in Section 3.2.2, except for not having pretrained SSL features. We also trained this model on LibriSpeech.

*3.2.2. Self-supervised speech representations*

We adopted the SSL configuration of ML-SUPERB, starting with a weighted sum of frozen SSL representations, where the weights are learnable, then applied SpecAugment and a convolutional downsampling layer reducing the sequence by half. The encoder consisted of 2 Transformer layers with 8 attention heads, a feed-forward dimension of 1024, and an output dimension of 256. The decoder consisted of 6 Transformer layers with 8 attention heads and a feed-forward dimension of 2,048. We trained the models with the Connectionist Temporal Classification (CTC) loss with a dropout rate of 0.1, and we used the Adam optimizer with a learning rate of 0.001. During decoding, we used beam search with a beam size of 60 and a CTC weight of 0.3. The model uses subwords learned from a unigram-based subword segmentation algorithm [41] as tokens and thus predicts subwords. (Words are recovered from the predicted subwords during evaluation). We use sentencepiece's unigram tokenization with a vocabulary size of 5,000.

We do not use language models, which in [9] yielded higher perplexity when a copula was omitted—a morphosyntactic feature of AAVE. Given the lack of such features in common speech corpora [22], language modeling would provide another source of discrepancy between MAE and AAVE, whereas our focus is on SSLR as features. However, we acknowledge that end-to-end ASR implicitly does language modeling.

We choose four popular pretrained SSLR models from SUPERB with 317 million parameters: `wav2vec2_large_960`, `hubert_large_ll60k`, `wavlm_large`, and `xls_r_300m`. We evaluate the models using word error rate (WER). Since WER does not assign partial credit to predictions that are phonetically similar to the ground truth [42], we use CER (character error rate) as well.

*3.2.3. Self-supervised speech pre-training data*

While wav2vec 2.0 and HuBERT are pre-trained on datasets primarily featuring read speech, WavLM and XLS-R are pre-trained on datasets encompassing diverse content sources, including spontaneous speech and different languages. Wav2vec 2.0 is pre-trained on Librispeech, which consists of read speech favoring MAE (Section 3). HuBERT leverages LibriLight-60k, which contains read speech from LibriVox but lacks demographic information. WavLM uses GigaSpeech, a corpus covering audiobooks, podcasts, and YouTube videos, potentially containing AAVE due to the diverse nature of YouTube. WavLM also uses VoxPopuli (En), which comprises 24k hours of English data from European Parliament recordings without explicit speaker information. XLS-R is pre-trained on multilingual datasets, with 372k hours across 23 languages from VoxPopuli-400k, 50k hours across 8 languages of LibriVox audiobooks from Multilingual LibriSpeech (MLS), 7k hours of read speech across 60 languages from CommonVoice (CV, v6.1) [5], 6.6k hours of YouTube across 107 languages from VoxLingua107 (VL), and 1K hours of conversational telephone speech across 17 African and Asian languages from Babel (BL).

### 3.3. Analysis

*3.3.1. Propensity score matching*

Following [9], we pair utterances across the two datasets using propensity-score matching. This alleviates the confounding effect that speakers' age and gender and clip duration can have on ASR performance by ensuring that their distribution across both datasets are similar. Using PsmPy [43], we train logistic regression models using sklearn that predict race given the speaker's age, speaker's gender, and the clip duration as features. We then use the logit as a propensity score and use nearest neighbor matching to match utterances across the datasets without replacement. After matching, we end up with 922 utterances in each dataset (Table 1).

*3.3.2. Dialect density measure (DDM)*

To quantify how much the use of AAVE features affects ASR performance, [9] scored each utterance with dialect density measure (DDM), which is the proportion of words in the utterance that demonstrated phonological and morphosyntactic features of AAVE. They manually annotated 150 utterances to obtain counts of these features, with 3.5 phonological and 0.5 grammatical features per utterance on average. Building on their methods, we re-ran inference just on these 150 utterances.[6] We then computed the correlation between the WER of an utterance and the DDMs provided by [9] to see whether or not SSLs struggle more with utterances with more AAVE features.

Table 2: *The performance of the four self-supervised speech representations on the matched CORAAL and NSP samples, in addition to the Pearson's correlation between AAVE Dialect Density Measure and WER on CORAAL. The two-sided p-value of each correlation was smaller than 1.96e-08. We also report the slope of the regression line where DDM is x and WER is y. LS WER refers to the performance of the model on the LibriSpeech test-clean subset.*

| Model | WER | | | CER | | | DDM-WER | |
|---|---|---|---|---|---|---|---|---|
| | LS | CORAAL | NSP | | CORAAL | NSP | Correlation | Slope |
| wav2vec 2.0 | 5.8 | 52.8 | 37.2 | | 33.1 | 21.7 | 0.505 | 1.008 |
| HuBERT | 4.0 | 43.7 | 29.8 | | 28.4 | 17.9 | 0.439 | 0.783 |
| WavLM | 3.4 | 28.1 | 21.3 | | 16.8 | 11.9 | 0.613 | 0.949 |
| XLS-R | 6.8 | 52.7 | 39.0 | | 32.8 | 23.0 | 0.592 | 1.080 |
| FBANK | 18.2 | 74.8 | 65.3 | | 51.1 | 42.9 | 0.463 | 0.731 |

## 4. Results and Discussion

### 4.1. SSLRs perpetuate the AAVE-MAE disparity

Table 2 shows that the WER of the SSL models is significantly higher on CORAAL than on NSP, indicating that SSL perpetuates the WER disparity between AAVE and MAE. The inter-dataset difference in WER (12.5) [7] is greater than the largest intra-dataset difference among regions (6.2 for CORAAL and 7 for NSP) [8], showing that the differences between MAE and AAVE affected our SSLRs more than the internal regional variation within the two datasets.

---

[5]Breakdowns by variety within the US are unavailable. Instead, the country of origin for non-native English speakers (L2) is reported.

[6]108 of the 150 utterances did not get matched to NSP utterances during propensity score matching and thus do not contribute to the WER of CORAAL in Table 2.

### 4.2. SSLRs struggle with linguistic features of AAVE

Not only do SSL speech models maintain the disparity between AAVE and MAE on the corpus level, the models also specifically struggle with phonological and morphosyntactic features of AAVE. Table 2 shows that for all four models, the Pearson's correlation between the dialect density measure and the WER is moderate. While the correlations are each lower than that of the proprietary models in [9] (0.7442), the persistence of the correlation (and the positive slope) means that SSL models still have higher WER on utterances with more features of AAVE.

A manual inspection of the most common mistakes made by SSL on CORAAL with jiwer [9] revealed many phonological and morphosyntactic features of AAVE that [9] considered when annotating DDM. [10] Across the four models, we identified final consonant cluster deletion ("it's" → "is" 461 times and "and" → "an" 207 times) and syllable initial fricative stopping ("that" → "dat" 69 times and "the" → "de" 46 times)[11]. At the subword level, we observed the suffix[12] "-s" inserted 1268 times in CORAAL (as opposed to 711 times in NSP), which demonstrates the absence of the third person singular present tense -s, possessive -s, or plural -s. Velar nasal fronting (the suffix "-in" instead of the suffix "-ing") occurs 90 times, though this feature is not specific to AAVE. The suffix "-t" is deleted 579 times in CORAAL but 362 times in NSP, the former of which could be neutralization of word-final /t/ [44]. At the phoneme level [13], we observed numerous deletions ("-t" 9035 times, "-d" 5853 times) and the stopping of voiceless interdental fricatives ("ð" → "d" 823 times). Additionally, there were multiple instances of final consonant cluster deletion, e.g. "nd" → "n" 876 times (14.42%), "nt" → "n" 506 times (8.33%), across 4 SSL models. [14] The FBANK model also exhibited similar errors on AAVE features but at a slightly lower proportion, e.g. "nd" → "n" 306 times (10.05%), "nt" → "n" 202 times (6.63%). This affirmed our previous findings that SSL pre-training reinforces the correlation between AAVE features and WER. When faced with phonological variation likely unseen during pre-training or training, SSL approximated the pronunciation instead of mapping the pronunciation variant to its conventional spelling. As for morphosyntactic variation, CORAAL preserved such features in its transcription, but the ASR is normalizing the speech towards MAE.

In short, SSL pre-training per se may not generalize to all features of AAVE during zero-shot ASR inference, suggesting explicit supervision with AAVE data may be needed.

### 4.3. SSL pre-training may not equally benefit all utterances

Consistent with [33], pretrained SSLRs perform better than the FBANK model with no pre-training (Section 3.2.1) across both CORAAL and NSP: for CORAAL, the WER of the FBANK model is 74.8 but decreases to 44.3 on average across the four SSL models; for NSP, 65.3 down to 31.8. While absolute improvement is similar (30.5 for CORAAL and 33.5 for NSP), the relative improvement is 40.8% for CORAAL and 51.3% for NSP. Additionally, the DDM-WER correlation was quite low for the FBANK model. That SSL increases the correlation in 3 of 4 cases suggests that SSL pre-training may not benefit all utterances of AAVE equally but can be biased against utterances with more features of AAVE.

## 5. Conclusion and Future Work

In our zero-shot ASR setting, pretrained SSLRs fail to bridge the AAVE-MAE disparity, struggle with features of AAVE, and are biased against utterances with more features of AAVE. While pre-training SSLRs on Mainstream American English reduces the amount of labeled data needed to reach lower WERs on MAE, current techniques alone cannot mitigate bias despite encoding phonetic information and capturing dialectal variation. Moving forward, pre-training or training corpora should contain AAVE [33], or we should employ other bias mitigation techniques such as data augmentation or more generalizable training objectives.

In the future, we will probe the SSL features to identify where they struggle with the phonological features of AAVE. We will match speakers across datasets with speaker embeddings or other neural alignment models. With computational resources, we would run controlled experiments with pre-training configurations (data, objective, and architecture) to attribute performance differences to differences in configuration [45].

## 6. Limitations

We evaluated on spontaneous speech but use models finetuned on read speech from LibriSpeech. Spontaneous speech is the more natural use case of ASR models. However, corpora of spontaneous speech with transcriptions on the scale of LibriSpeech are hard to find. We attempted matched n-grams [34] to control for the differences in text content between NSP and CORAAL, but the overlap between matched utterances comprises only 250 bigrams and 39 trigrams (omitting filler words). While we reduced the effect of age, gender, and clip duration via propensity score matching, the recording quality and differences in transcription practices between the two corpora may also contribute to WER. The WER on NSP may also be affected by the quality of the forced alignment-based segmentation, as the timestamps for conversational speech may not always be accurate.

## 7. Acknowledgement

We thank Shinji Watanabe for providing compute and feedback on our experimental setup and methodology. We used the Bridges2 system at PSC through allocation CIS210027P from the Advanced Cyberinfrastructure Coordination Ecosystem: Services & Support (ACCESS) program. We also thank Jeremiah Milbauer for his thorough feedback across iterations of an earlier draft of our work. We finally thank May Pik Yu Chan for her Praat script on extracting vowel formants.

---

[7] Here we concatenate the 4 SSLRs' predictions for each dataset.
[8] 34.5 (Mid-Atlantic) - 27.5 (Midland) for NSP and 46.3 (Princeville) - 40.1 (Rochester) for CORAAL
[9] https://github.com/jitsi/jiwer
[10] We used jiwer only to obtain the most common edits; all evaluation metrics (including DDM-WER correlation) use sclite.
[11] Reference → hypothesis
[12] Sentencepiece subwords without preceding underscores (spaces) are suffixes, not necessarily morphologically but in the string sense.
[13] We obtained phonemes for both reference and hypothesis with https://github.com/Kyubyong/g2p.
[14] However, when it was unclear whether the process was phonological or morphosyntactic (e.g. *named* → *name*), transcribers followed standard orthographical conventions.